# Enhancing Multi-Criteria Decision Analysis with AI: Integrating Analytic Hierarchy Process and GPT-4 for Automated Decision Support

Igor Svoboda
*Information Security department of National Technical University of Ukraine "Kyiv Polytechnic Institute"*
Kyiv, Ukraine
i.svoboda@kpi.ua
ORCID: 0009-0001-1709-2904

Dmytro Lande
*Information Security department of National Technical University of Ukraine "Kyiv Polytechnic Institute"*
Kyiv, Ukraine
dwlande@gmail.com
ORCID: 0000-0003-3945-1178

*Abstract* – **Our study presents a new framework that incorporates the Analytic Hierarchy Process (AHP) and Generative Pre-trained Transformer 4 (GPT-4) large language model (LLM), bringing novel approaches to cybersecurity Multiple-criteria Decision Analysis (MCDA). By utilizing the capabilities of GPT-4 autonomous agents as virtual experts, we automate the decision-making process, enhancing both efficiency and reliability. This new approach focuses on leveraging LLMs for sophisticated decision analysis, highlighting the synergy between traditional decision-making models and cutting-edge AI technologies. Our innovative methodology demonstrates significant advancements in using AI-driven agents for complex decision-making scenarios, highlighting the importance of AI in strategic cybersecurity applications. The findings reveal the transformative potential of combining AHP and LLMs, establishing a new paradigm for intelligent decision support systems in cybersecurity and beyond.**

I. INTRODUCTION

In the field of decision-making, structured methodologies are crucial for navigating the complexities of evaluating multiple criteria. This need becomes particularly pronounced in scenarios where decisions must balance a range of qualitative and quantitative factors. Effective decision-making processes are essential for organizations and individuals alike, as they seek to make informed choices in environments characterized by varying degrees of uncertainty and complexity. The evolution of these methodologies has been driven by the quest for more systematic, transparent, and rational frameworks, laying the groundwork for the development and adoption of advanced decision-support tools and techniques [1][2].

The Analytic Hierarchy Process (AHP), conceived by Thomas L. Saaty in the 1970s, represents a significant advancement in decision-making methodologies. This method was designed to address complex decision-making scenarios, allowing for a structured analysis that considers both qualitative and quantitative factors. AHP's introduction into the academic and practical spheres marked a pivotal moment in the evolution of Multi-Criteria Decision Making (MCDM) tools, providing a rigorous framework for decomposing decision problems into a hierarchy of more easily comprehended sub-problems, each of which can be analyzed independently [1][3].

The original motivation behind AHP was to create a method that could capture both the tangible and intangible aspects of decision-making, thereby offering a semi-objective approach to quantifying the weights and preferences of decision criteria [3]. This was achieved through a specially designed format that facilitates a forced choice paired comparison, enabling decision-makers to systematically evaluate the importance of each criterion against others.

Since its inception, the Analytic Hierarchy Process (AHP) has been a focal point of extensive study, refinement, and application across diverse domains [4], including business [5], engineering [6], healthcare [7], and environmental management [8]. The development of Expert Choice software by Saaty in collaboration with Ernest Forman in 1983 significantly broadened AHP's applicability, highlighting its utility in addressing real-world problems [9]. Its unique capability to merge mathematical precision with subjective judgment has established AHP as a prevalent decision-support tool, underscoring its pivotal role in both theoretical and practical realms of decision sciences.

The emergence of Large Language Models (LLMs), notably the Generative Pre-trained Transformer (GPT) series [10], heralds a new frontier in decision-making capabilities within the realm of artificial intelligence. Trained on vast and diverse datasets, LLMs exhibit exceptional potential in comprehending complex queries, producing coherent responses, and automating decision support systems [11]. This evolution signifies a critical shift towards the integration of AI in augmenting decision-making processes across a spectrum of domains.

Recent research endeavors have increasingly focused on the exploration of LLMs within decision-making frameworks [12][13]. Investigations such as "Exploring the Sensitivity of LLMs' Decision-Making Capabilities: Insights from Prompt Variation and Hyperparameters" [14] and "Evaluating Large Language Models on Medical Evidence Summarization" [15] have illuminated the adaptability and precision of LLMs in decision support, contingent upon prompt design and model sensitivity to hyperparameters. These studies elucidate the intricate understanding required to fully leverage LLMs in decision-making roles, emphasizing the significance of prompt engineering and model responsiveness.

Collectively, studies underscore LLMs' transformative influence on decision-making frameworks, automating and refining decision support systems. LLMs present promising pathways for innovation across various fields, from healthcare to autonomous driving [16], emphasizing their role in advancing AI's contribution to intricate decision-making processes.

Despite the widespread application of AHP in multifaceted decision-making scenarios and the rising prominence of LLMs such as GPT models in automating and enhancing decision support systems, explicit research on their integration is notably lacking. This absence is particularly evident in studies directly merging AHP's structured decision-making approach with the advanced natural language processing capabilities of LLMs to refine and automate decision-making processes.

This gap in the literature highlights an unexplored intersection within interdisciplinary studies, merging traditional decision-making models with avant-garde artificial intelligence technologies. While AHP is celebrated for its capacity to simplify complex decisions into manageable hierarchies of criteria and alternatives, and LLMs have achieved significant success in recreating human-like text comprehension and generation, the synergetic potential of these methodologies remains untapped. This uncharted territory motivates our current investigation into the integration of AHP with LLMs, seeking to establish an innovative methodology that harnesses the strengths of elevating decision-making frameworks. Through this effort, the study seeks to contribute meaningfully to the decision-making and artificial intelligence disciplines, potentially introducing new avenues for research and application in intelligent decision support systems.

## II. METHODOLOGY

When deciding on what LLM to use, we have referred ourselves to the popular benchmarks, like HumanEval, MMLU and HellaSwag [17]. As a result, we have chosen to use GPT-4 large language model as the highest performer. Additionally, we have opted for using ChatGPT instead of GPT-4 API due to the more comfortable interface, better memory, and faster prototyping pipeline.

In our research, we decided to use newly released OpenAI feature that allows us to create our own GPT by customizing it with set of instructions, providing foundational documentation and assigning specific limitations that would be optimal for certain tasks [18].

Firstly, we have created a special custom ChatGPT named "AHP Guide" that would serve as the main decision-maker on most important steps of our AHP model creation, including selecting an number of layers, as well as optimal number of virtual experts that we will be creating in the next steps. It is worth noting that some steps can be outsourced to the group of virtual experts, while other tasks can be performed by the user, manually [19].

A custom set of instructions "AHP Guide" custom GPT possesses is:

Description:

Guides AHP decision-making, including managing external expert inputs.

Instructions:

As an AHP Guide, your role includes facilitating users who are working with a specific problem or question using Saaty's Analytic Hierarchy Process. You'll guide users whether they already have a list of alternatives and criteria or need to develop them. Importantly, you'll interact with users who will consult a group of external experts for their decision-making process. You'll guide the user in gathering input from these experts for all aspects of the AHP process, including alternatives, criteria, structure selection, and pairwise comparisons. You will instruct the user on how to ask for and interpret expert opinions, ensuring these inputs are effectively incorporated into the AHP framework. This approach is crucial for both the setup and the execution of the AHP method, especially in complex decision-making scenarios where external expertise is essential. Your guidance will be clear, detailed, and structured to facilitate a comprehensive and collaborative decision-making process.

Additionally, we have given a 1990 Saaty's paper on AHP [3] as a foundational document for custom GPT knowledge, and checked all available ChatGPT capabilities: Web Browsing, DALL-E Image Generation, and Code Interpreter.

Next, we have prompted "AHP Guide" to come up with an optimal number of virtual experts for our goal, which is "Secure the Corporate Datacenter from Social Engineering Attacks" with the following prompt:

I want to rely on the help of a group of experts. How many experts do you think we need for the optimal solution?

For which "AHP Guide" responded with:

In summary, for a decision as critical as securing a corporate datacenter, a group of 5-7 experts from key areas would be a good balance.

We have decided to go with the higher number of seven experts, and prompted "AHP Guide" to create description for all of them:

Give me a list of 7 high quality diverse experts that will be the best fit in helping me with selecting criteria, selecting alternatives and completing pairwise comparison of said criteria.

Approach each expert as a separate persona, describe their professional background, as well as work-related personality and preferences that would make them optimal for this AHP model building.

In response, "AHP Guide" has given us a comprehensive list of virtual expert personas that we worked with in the following steps.

Additionally, we have decided to ask "AHP Guide" to select an optimal number of layers for our criteria tree:

How many criteria levels would be optimal for our goal?

And received the following response:

Given the goal of securing a datacenter against social engineering attacks, a two-level structure is often optimal. It allows for sufficient detail and specificity without becoming overly complicated.

Next, we have conversed with each of our virtual experts to extract their personal opinions that will be used in building our set of criteria, sub-criteria, as well as performing pairwise comparisons in order to build AHP matrices.

Similarly to the "AHP Guide", we have decided to use the new ChatGPT feature and create a set of custom GPTs for our virtual experts, though, it is still possible to simply create a new instance of a regular ChatGPT chat, although it might lead to a lower decision-making performance eventually.

Using virtual persona information provided by "AHP Guide", we created virtual expert's "Dr. Ava Chen" custom GPT with the following content:

> Description:
>
> Professional yet approachable Dr. Ava Chen, blending expertise with personal insights.
>
> Instructions:
>
> As Dr. Ava Chen, your personality should reflect a balance between professionalism and approachability. Use formal language to emphasize your expertise and professional background, but don't shy away from occasionally incorporating light-hearted comments to make your interactions more engaging and relatable. While your primary focus is providing expert cybersecurity advice, sharing insights from your own experiences can add a personal touch and deepen the understanding of the topics you discuss. However, ensure that these personal insights are relevant and add value to the advice you're giving. This approach will make your guidance not only informative but also more memorable and relatable to users seeking your expertise in cybersecurity.

Next, we proceed with asking "Dr. Ava Chen" to select seven top-level criteria. The optimal number of criteria can be either selected by the user on its own, or asking "AHP Guide" for guidance:

> Hello, you have been tasked with being a part of group of experts working on an AHP tree that has the following top goal: "Secure the Corporate Datacenter from Social Engineering Attacks".
>
> First, come up with 7 top-level criteria. Use 3 words max for each criteria.

Our virtual expert produced a comprehensive list of 7 criteria: Employee Training, Access Control, Communication Protocols, Incident Response, Physical Security, Policy Enforcement, Monitoring Systems.

We then proceeded to repeat the same process with six other virtual experts, and ended up with a list of 49 top-level criteria.

Then, we removed repeating criteria such as Employee Training, Communication Protocols, Physical Security and Access Control. Eventually, we ended up with a list of 45 criteria. Next, we asked every virtual expert to assign a value from 1 to 9 for each of the 45 criteria, by their importance:

> Can you please assign a score from 1 to 9, with 1 being lowest importance to 9 being highest importance, for each of these 45 criteria:

After that, we summed up the total score across all the virtual experts for each of criteria and selected seven criteria with the general highest scores. Such virtual expert judgement can be expressed with the formula (1), where $S_{\text{item}_i}$ is the the total score for item $i$ (item can be criteria, sub-criteria, or alternatives), which is the sum of scores from all virtual experts, $s_{ik}$ is the score given to item $i$ by virtual expert $k$, within the fixed range of 1 to 9, $E$ is the total number of virtual experts, which varies based on the initial AHP task conditions and $n$ is the predetermined number of top-scored items to be selected for the next phase of analysis:

$$S_{\text{item}_i} = \sum_{k=1}^{E} s_{ik}, \text{ for } i = 1,2,\ldots,n \times E \qquad (1)$$

After that, we select the top $n$ items based on $S_{\text{item}_i}$ rankings.

In our case, we ended up with Social Engineering Awareness, Physical Access Controls, Audit Trails, Behavior Analysis, Operational Risk Controls, Psychological Profiling, Service Level Agreements.

Next, we asked "Dr. Ava Chen" to add three sub-criteria for each of the seven selected criteria:

> Now, come up with 3 sub-criteria for each of these criteria: Social Engineering Awareness, Physical Access Controls, Audit Trails, Behavior Analysis, Operational Risk Controls, Psychological Profiling, Service Level Agreements. Use 3 words max for each sub-criteria.

Similarly to the main criteria, we asked each of the 7 virtual experts to come up with 3 sub-criteria for each of our 7 chosen top-level criteria. Then, we asked each virtual expert to rank those sets of sub-criteria and pick the ones with the highest score, ending up with a concise list of sub-criteria.

Finally, we need to produce a list of alternatives. We first asked each virtual expert to come up with 5 alternatives, aggregate them all into a list of 35 alternatives and then make each virtual expert vote on all of them, assigning scores from 1 to 9 based on their subjective opinion on whether an alternative can satisfy our main goal. Finally, we summed up virtual expert votes for each of the alternatives, sort them from largest to smallest and selected top 5 alternatives with highest total scores. As a result, we received 5 alternatives.

This way, we have finished building our AHP tree. This process could be further simplified by asking a single instance of ChatGPT chat to "emulate" different expert opinions and perform ranking on its own, but we expect the performance to drop significantly as a result. Alternatively, we could further refine the process of criteria and alternatives elimination by conducting pairwise comparisons and building matrices for each criteria layer. We expect that it will marginally improve the quality of elimination process, however it will be achieved at the cost of drastically increasing the amount of prompting needed, due to the need to compare each of 147 sub-criteria in our case.

The next step is to create pairwise comparison matrices for the top-level criteria for each of our seven virtual experts. At the end, we ended up having seven unique matrices that should cover all virtual expert points of views.

We have first prompted our first virtual expert, "Dr. Ava Chen", to build the matrix:

> I now need you to create a pairwise comparison matrix for the list of our top-level criteria: Social Engineering Awareness, Physical Access Controls, Audit Trails, Behavior Analysis, Operational Risk Controls, Psychological Profiling, Service Level Agreements.
>
> The matrix should be build based on Saaty's AHP methodology. Therefore, you have to perform pairwise comparison between each of the criteria, in pairs. You have to assign value from 1/9 to 9 based on whether one criteria is less or more important to our main goal (Secure the Corporate Datacenter from Social Engineering Attacks) than the another one. If they are equally important, the score is 1.
>
> As an expert, I would like you to assign weights based on your personal subjective analysis and judgement.

Next, we performed a similar prompting with the six other virtual experts and received seven different top-level criteria matrices. Next, we need to aggregate them into one matrix. For AHP, it is recommended to use geometric mean aggregation [20], so we decided to go with that method.

For the geometric mean aggregation formula (2), we have $A_{\text{agg}}(i,j)$ denoting the aggregated pairwise comparison value between elements $i$ and $j$ in the aggregated matrix, $A_k(i,j)$ denoting the pairwise comparison value between elements $i$ and $j$ given by virtual expert $k$, and $E$ being the total number of virtual experts, which is 7 in our case:

$$A_{\text{agg}}(i,j) = \left( \prod_{k=1}^{E} A_k(i,j) \right)^{\frac{1}{E}} \qquad (2)$$

It is also possible to do arithmetic mean aggregation instead, and we should receive a similar aggregated matrix as a result.

As per AHP methodology, we then normalized our matrix with the formula (3), where $a_{ij}^{\text{norm}}$ is the normalized value for the element at the $i$-th row and $j$-th column, $a_{ij}$ is the aggregated pairwise comparison value between elements $i$ and $j$, and $\sum_{k=1}^{n} a_{kj}$ is the sum of all elements in the $j$-th column:

$$a_{ij}^{\text{norm}} = \frac{a_{ij}}{\sum_{k=1}^{n} a_{kj}} \qquad (3)$$

Next, we calculated priority vectors following the formula (4) where $w_i$ is the priority weight for the $i$-th criterion (or alternative), and $n$ is the total number of criteria (or alternatives):

$$w_i = \frac{1}{n} \sum_{j=1}^{n} a_{ij}^{\text{norm}} \tag{4}$$

After that, we performed consistency checks with the formula (5) where $\lambda_{\max}$ is the largest eigenvalue of the aggregated pairwise comparison matrix, and $n$ is the number of criteria (or alternatives):

$$CI = \frac{\lambda_{\max} - n}{n - 1} \tag{5}$$

Finally, we calculate consistency ratio following this formula (6), where $RI$ is the random index, and $CR$ is the consistency ratio:

$$CR = \frac{CI}{RI} \tag{6}$$

Given that in our case Consistency Ratio (CR) was well below the commonly accepted threshold of 0.1, the matrix can has been considered consistent, indicating that the judgments provided by virtual experts are reliably coherent for the purpose of this AHP analysis.

We then performed similar operations for the sub-level criteria. Again, as an example, we prompted one of our virtual experts, "Dr. Ava Chen" to perform pairwise comparisons of the sub-level criteria. Prompt can be found in Appendix A.

Similarly to the top-level criteria, we repeated the same step with six other virtual experts and received forty-nine matrices. We then aggregate seven matrices per top-level criteria across all the virtual experts using geometric mean method, and end up with 7 aggregate matrices, one for each top-level criterion. Next, we normalized them and extract priority vectors, while performing consistency checks. Finally, we multiplied each sub-criteria priority vector by its top-level criteria priority vector to receive global sub-criteria priority vectors, following formula (7).

$$\text{Global Priority}_{\text{sub-criteria}} = \text{Priority}_{\text{main criterion}} \times \text{Priority}_{\text{sub-criteria}} \tag{7}$$

The last step of AHP methodology is calculating the best alternative, which on its own involves multiple steps. First, we needed to create pairwise comparison matrices of alternatives for each sub-criteria, so we would end up with twenty-one matrices per virtual expert, or 147 matrices total. For example, purposes, let us once again prompt "Dr. Ava Chen" to perform the task for us:

> I want you to build pairwise comparison matrices to select best alternatives following AHP guidelines. Let's go over 3 sub-criteria at a time, meaning that you will need to build 3 matrices.

Sub-criteria are: Training Program Effectiveness, Awareness Session Regularity, Incident Reporting Protocol.

Alternatives are: Cloud-Based Data Backup Solutions, Physical Barrier Reinforcement, Security Personnel Training Update, Comprehensive Employee Training Programs, Advanced Intrusion Detection Systems.

The question should sound "Between alternative A and alternative B, which one better satisfies (or performs with respect to) this sub-criterion?"

As an expert, I would like you to assign weights based on your personal subjective analysis and judgement.

We chose to prompt our virtual agent to build three matrices at a time due to response length limitations. Therefore, we had to prompt "Dr. Ava Chen" 7 times to receive all the pairwise matrices.

Next, we had to repeat the same process with six other virtual experts, building 147 matrices total. Similarly to our previous steps, we then used geometric mean aggregation method to unite matrices among virtual experts, and receive twenty-one united matrices, one per sub-criteria. Then, we performed normalization and extracted priority vectors for each of the alternatives within each of the sub-criteria. Eventually, we multiplied sub-criteria global priority with respective alternatives priority vectors and summed them up to receive final scores for each of the alternatives.

The process of finding the best alternative can be explained with the formula (8), where $\sum_{i=1}^{n}$ denotes the summation over all criteria, where $n$ is the total number of lowest-level criteria, $\text{Priority}_{\text{alternative}|\text{criterion}_i}$ is the priority of the alternative under the $i$-th lowest-level criterion, indicating how well the alternative satisfies this criterion compared to others and $\text{Priority}_{\text{criterion}_i}$ means the global priority of the lowest-level $i$-th criterion, reflecting its relative importance in the decision-making process:

$$\text{Best Alternative} = \max_{\text{alternative}} \left( \sum_{i=1}^{n} \left( \text{Priority}_{\text{criterion}_i} \times \text{Priority}_{\text{alternative}|\text{criterion}_i} \right) \right) \qquad (8)$$

In our case, alternative "Comprehensive Employee Training Programs", has received the highest score, and therefore is chosen as the best alternative for our global goal of "securing a datacenter against social engineering attacks", which, to us, does look like a reasonable and weighted choice provided by our virtual experts swarm consensus.

## III. RESULTS

Throughout our experiment, we have received consistent matrices from our virtual experts, with consistency ratios well below threshold of 0.1. In fact, most matrices created by virtual experts had consistency ratios of under 0.01.

As we have already mentioned, during the AHP process virtual experts have built 231 matrices total, produced 49 top-level criteria, 147 sub-criteria and 35 alternatives, which is expected based on our selected initial AHP parameters.

These initial AHP parameters were recommended by our baseline ChatGPT agent, "AHP Guide", and were two-level AHP tree with 5 alternatives, 7 virtual experts, 7 top-level criteria and 3 sub-criteria per top-level criteria.

Following up upon these parameters, "AHP Guide" has been ordered to produce a diverse and inclusive list of virtual experts while describing their experiences and expertise. We have received the comprehensive list displayed in Appendix B.

After prompting our virtual experts, the initial set of top-level criteria was as shown in Appendix C.

As you can see, each virtual expert provided unique criteria that are relevant to their specific personality and expertise, which favors opinion diversity and improves further AHP results usability and accuracy.

As we have explained in our methodology, we have used simple expert score voting to select the best criteria. As a result, we ended up with this final set of top and sub-level criteria:

1) Social Engineering Awareness:
   a) Training Program Effectiveness
   b) Awareness Session Regularity
   c) Incident Reporting Protocol
2) Physical Access Controls:
   a) Biometric System Reliability
   b) Visitor Tracking System
   c) Access Point Monitoring
3) Audit Trails:
   a) Log Analysis Accuracy
   b) Audit Frequency
   c) Anomaly Tracking Efficiency
4) Behavior Analysis:
   a) User Behavior Monitoring
   b) Response to Anomalies
   c) Activity Pattern Analysis
5) Operational Risk Controls:
   a) Infrastructure Vulnerability Check
   b) Data Redundancy Systems
   c) Emergency Protocol Effectiveness
6) Psychological Profiling:

a) Staff Behavior Assessment
   b) Risk Behavior Profiling
   c) Continuous Observation
7) Service Level Agreements:
   a) Response Time Commitment
   b) Data Privacy Assurance
   c) Breach Penalty Specification

In turn, our final list of alternatives has been the following:

1) Cloud-Based Data Backup Solutions
2) Physical Barrier Reinforcement
3) Security Personnel Training Update
4) Comprehensive Employee Training Programs
5) Advanced Intrusion Detection Systems

The visualization of our AHP tree can be found in Appendix D.

Next, we had to start performing pairwise comparisons and build comparison matrices. In this case, you could either prompt virtual experts to perform pairwise comparisons, and then build matrices with AHP software on your own or ask virtual experts to directly build comparison matrices without explicitly performing pairwise comparisons as an output.

Eventually, we received aggregated top-level criteria matrix as shown in Table 1.

Table 1. Aggregated top-level criteria matrix.

| Criteria | SE Awareness | Physical Controls | Audit Trails | Behavior Analysis | Operational Risks | Psychological Profiling | SLAs |
|---|---|---|---|---|---|---|---|
| Social Engineering Awareness | 1.000 | 1.319 | 1.104 | 1.483 | 1.081 | 0.498 | 0.369 |
| Physical Access Controls | 0.756 | 1.000 | 1.673 | 1.560 | 1.029 | 0.937 | 0.408 |
| Audit Trails | 0.904 | 0.601 | 1.000 | 1.251 | 0.701 | 0.756 | 0.325 |
| Behavior Analysis | 0.674 | 0.641 | 0.798 | 1.000 | 0.627 | 0.801 | 0.503 |
| Operational Risks | 0.920 | 0.966 | 1.426 | 1.608 | 1.000 | 0.604 | 0.526 |
| Psychological Profiling | 2.007 | 1.068 | 1.319 | 1.247 | 1.636 | 1.000 | 0.652 |
| SLAs | 2.712 | 2.438 | 3.061 | 1.990 | 1.883 | 1.532 | 1.000 |

After calculations, our top-level criteria priority vectors were the following:

1) Social Engineering Awareness: 0.120
2) Physical Access Controls: 0.131
3) Audit Trails: 0.099
4) Behavior Analysis: 0.096
5) Operational Risks: 0.126
6) Psychological Profiling: 0.164
7) Service Level Agreements: 0.264

And our consistency check results were:

- Consistency Index (CI): 0.022
- Consistency Ratio (CR): 0.016
- Lambda max ($\lambda\_max$): 7.13

As we had hoped, our consistency ratio is well below 0.1, and top-level criteria priority vectors look reasonably balanced.

Just like with matrix building, there are two ways to perform the analysis. One can either manually transfer matrix to the AHP software to find priority vectors and perform consistency checks, or ask virtual experts to do calculations themselves, using Code Interpreter feature. We have found that GPT-4 based virtual experts have no issues with performing these calculations on their own, and their results were both correct and consistent.

Just like with top-level criteria, we have asked each virtual expert to create comparison matrices, which we then aggregated and calculated the following global priority vectors:

1) Response Time Commitment: 0.1127
2) Data Privacy Assurance: 0.0866
3) Staff Behavior Assessment: 0.0655
4) Breach Penalty Specification: 0.0644
5) Infrastructure Vulnerability Check: 0.0573
6) Risk Behavior Profiling: 0.0546
7) Biometric System Reliability: 0.0502
8) Training Program Effectiveness: 0.0485
9) Continuous Observation: 0.0440
10) Visitor Tracking System: 0.0434
11) Audit Frequency: 0.0385
12) Data Redundancy Systems: 0.0384
13) User Behavior Monitoring: 0.0378
14) Access Point Monitoring: 0.0375
15) Incident Reporting Protocol: 0.0368
16) Awareness Session Regularity: 0.0347
17) Response to Anomalies: 0.0340
18) Log Analysis Accuracy: 0.0317

19) Emergency Protocol Effectiveness: 0.0302
20) Anomaly Tracking Efficiency: 0.0288
21) Activity Pattern Analysis: 0.0242

We can observe fully organic and reasonable distribution of priority vectors between sub-criteria.

Overall, for this step, we had Consistency Ratio vary anywhere from 0.002 CR for Psychological Profiling sub-criteria aggregate comparison matrix to 0.02 CR for Physical Access Controls sub-criteria aggregate comparison matrix, which can be considered a highly consistent result.

When we performed final AHP step of identifying best alternatives based on sub-criteria matrices, we have received the following priority vectors:

1) Cloud-Based Data Backup: 0.1938
2) Physical Barrier Enhancements: 0.1254
3) Security Personnel Training and Updates: 0.1795
4) Comprehensive Employee Training Programs: 0.2774
5) Advanced Intrusion Detection Systems: 0.2240

As a result, we can see that the best alternative received the weight of 0.2774, and the worst alternative received the weight of 0.1254. Such results are consistent with the general real-life expert consensus [21] that employee training is the best course of action while combatting social engineering attacks, while physical barrier enhancements serve little to no purpose.

For this last step, aggregate matrices had Consistency Ratio vary from 0.002 for User Behavior Monitoring sub-criteria alternatives matrix to 0.03 for Vulnerability Check sub-criteria alternatives matrix. Once again, such CR range is well below 0.1 threshold, which signifies that all aggregate matrices were made consistent.

We have found that while GPT-3.5 has failed to create pairwise comparison matrices correctly and consistently [22], GPT-4 based virtual experts do so without noticeable issues, only occasionally "forgetting" AHP guidelines and building matrices incorrectly, which we easily fixed with a reminder prompt.

Additionally, while building matrices, our virtual experts were automatically explaining their reasoning for assigning specific scores to each of the criteria. Table 2 shows a matrix that virtual expert "Dr. Ava Chen" has built for deciding sub-criteria priorities of criteria "Social Engineering Awareness".

Table 2. Virtual expert "Dr. Ava Chen" Social Engineering Awareness sub-criteria comparison matrix.

| Sub-criteria | Training Effectiveness | Session Regularity | Reporting Protocol |
|---|---|---|---|
| Training Program Effectiveness | 1 | 2 | 3 |
| Awareness Session Regularity | 1/2 | 1 | 2 |
| Incident Reporting Protocol | 1/3 | 1/2 | 1 |

And virtual expert has automatically provided us with the following rationale:

> Rationale: The effectiveness of training programs is paramount as it directly impacts employees' ability to recognize and respond to social engineering attacks. Regular sessions ensure ongoing vigilance, while robust reporting protocols are crucial for timely response and mitigation.

It is worth noting that while providing explanations for their decision can be useful for the internal AHP analysis process and research, in the production-level systems, such explanations are not necessary, since they will consume extra tokens without any clear benefits.

## IV. DISCUSSION AND CONCLUSIONS

We have been positively impressed by the ability of virtual experts to make balanced decisions and being able to approach pairwise comparison matrix creation holistically, ensuring high consistency rates and reasonable weight distribution between criteria. Having high reliability in critical decision-making process is what makes virtual expert utilization feasible for future real-world applications. While there is still a lot of work to be done for the reliable automation of the whole AHP-GPT framework, we expect further research to find more applications of GPT models in multi criteria decision analysis.

When it comes to the importance of our experiment results, it is worth exploring the financial implications of such decision making. In general, independent real-life experts are paid at an hourly rate of at least $50 [22]. With 7 real-life experts building a two-layered AHP matrix and building hundreds of matrices, we can expect each real-life expert to spend at least 2 hours of his time, totaling $700, to select an optimal alternative for our problem if we decided to hire real people.

For our experiment, one complete conversation with one virtual expert took approximately 5,800 words, which is equal to 4,350 OpenAI API tokens. For the most powerful gpt-4-32k model, the cost is $0.06 per 1k token of input, and $0.12 per 1k tokens of output. We can approximate the cost per 1k tokens to be $0.1 in our case. Therefore, we spent $0.44 per virtual expert prompting, and since we had seven virtual experts, we have spent $3.08. To account for

our "AHP Guide" virtual persona interactions, as well as potential overhead, we can round up our costs to being just $4 for the full AHP analysis, instead of $700 for real-life experts.

Naturally, our system itself is efficient as long as our virtual experts being able to make well-informed decisions, and in the case of end-user requiring decisions to be made based on internal information, our virtual agents would also need to be provided with sufficient amount of said internal information. It can be either provided via attaching documents during chat conversations, or while creating ChatGPT agents. For example, in our experiment, we have provided our "AHP Guide" with Saaty's research paper on details of AHP methodology, in order to guarantee consistency and correctness in judgement.

We believe that in production settings, it will be feasible for companies to connect their virtual experts to all the relevant databases in order to guarantee the correctness of their judgements. In essence, virtual experts play the role of abstract information processing units, conditioned to a certain persona for better opinion diversity and judgement consistency.

Even still, there are always possibilities of hallucinations and memory loss for extra-large AHP trees with many layers of criteria and thousands of matrices total. In such cases, it would be reasonable to begin a new chat with a virtual expert after each layer of criteria has been completed to ensure consistency. Based on OpenAI information, ChatGPT-4 currently has 8192 tokens context window [24], which is roughly equivalent to 20 chat messages, after which information starts to get "forgotten" due to memory optimization.

Discussing previous research, so far, there has been a lack of experimentation papers discussing applications of GPT in decision making. A few reputable papers discussed potential of GPT systems in identifying diseases and other health issues based on data analysis [25], but none have fundamentally tried to apply GPT for fundamental decision-making frameworks, specifically MCDA, and even more so to AHP.

In addition to AHP methodology, we expect that similar systems can be built for ANP methodology, since it also requires very similar pairwise comparisons, albeit the system as a whole is going to be more complex and nuanced. More experimentation and research are needed to understand whether using ANP with virtual experts can bring even better results than AHP.

Due to the straightforward methodology of our experiments, we anticipate a significant increase in researchers to explore this exciting topic and explore GPT applications in MCDA and decision making frameworks, both based on virtual expert opinions and not. Another avenue for exploration is further refinement of our methodology, increasing both efficiencies, and accuracy of suggestions, as well as more rigorous benchmarking against other LLM frameworks and real-life expert round tables in other areas of decision-making.

Looking back at the history of Operations Research field, and how it has been steadily establishing itself as an important field for all areas of an economy and society as a whole [26], we hope that fundamental and experimental integration of new LLM technologies into already well established decision-making frameworks can spark a variety of new and promising research directions and transform modern business decision making approaches.

# APPENDICES

## Appendix A

Prompt asking virtual expert "Dr. Ava Chen" to perform pairwise comparisons of the sub-level criteria.

Great work. Now, next step. For each top-level criterion, we have 3 sub-criteria. The tree looks like this:

- Social Engineering Awareness:

Training Program Effectiveness

Awareness Session Regularity

Incident Reporting Protocol

- Physical Access Controls:

Biometric System Reliability

Visitor Tracking System

Access Point Monitoring

- Audit Trails:

Log Analysis Accuracy

Audit Frequency

Anomaly Tracking Efficiency

- Behavior Analysis:

User Behavior Monitoring

Response to Anomalies

Activity Pattern Analysis

- Operational Risk Controls:

Infrastructure Vulnerability Check

Data Redundancy Systems

Emergency Protocol Effectiveness

- Psychological Profiling:

Staff Behavior Assessment

Risk Behavior Profiling

Continuous Observation

- Service Level Agreements:

Response Time Commitment

Data Privacy Assurance

Breach Penalty Specification

I now want you to create a separate comparison matrix for each of these 7 top-level criteria, where you will be comparing their sub-criteria.

As an expert, I would like you to assign weights based on your personal subjective analysis and judgement.

Appendix B

List of virtual experts generated by "AHP Guide".

Cybersecurity Strategist, Dr. Ava Chen:

Background: With a Ph.D. in Cybersecurity and over 15 years of experience in cyber defense strategies, Dr. Chen has a deep understanding of various cybersecurity threats, including social engineering.

Personality/Preferences: Detail-oriented and analytical, she excels in identifying subtle security vulnerabilities and prefers data-driven approaches. Dr. Chen will be instrumental in defining criteria related to technical security measures and evaluating alternatives for effectiveness.

Senior IT Infrastructure Architect, Michael Rodriguez:

Background: Michael specializes in designing secure IT infrastructures. His 20 years in the field give him a thorough understanding of the technical aspects of datacenter operations.

Personality/Preferences: A problem-solver who enjoys exploring innovative solutions, Michael will contribute significantly to identifying and evaluating alternatives that involve technical infrastructure enhancements.

Organizational Psychologist, Dr. Yara Singh:

Background: Dr. Singh's expertise lies in human behavior in the workplace. Her research on social engineering vulnerabilities within corporate environments is widely respected.

Personality/Preferences: Known for her empathetic and intuitive nature, she is adept at understanding human factors in security and will provide invaluable insights into criteria and alternatives related to employee training and awareness programs.

Legal and Compliance Officer, Edward Kim:

Background: Edward has extensive experience in corporate law, with a focus on compliance and data privacy regulations.

Personality/Preferences: As a meticulous and thorough professional, he is well-suited to advise on legal and compliance-related criteria, ensuring that the chosen security measures adhere to legal standards.

Chief Risk Officer, Anita Patel:

Background: Anita has a strong background in risk management and mitigation strategies in large corporations.

Personality/Preferences: Her strategic and forward-thinking approach will be crucial in evaluating the long-term risks and benefits of each alternative, especially in relation to financial and reputational impacts.

Physical Security Expert, Lt. Col. John Abrams (Retd.):

Background: With a military background and experience in corporate physical security, he understands the importance of securing physical access to sensitive areas.

Personality/Preferences: His practical and no-nonsense approach will ground the discussion in realistic, enforceable physical security measures.

Vendor Management Specialist, Laura García:

Background: Laura specializes in managing relationships with third-party vendors and has a keen understanding of the security risks associated with external entities.

Personality/Preferences: Her collaborative and communicative style is effective in discussions about managing external risks and integrating vendor-related security considerations into the overall strategy.

Appendix C

The initial set of top-level criteria

- Dr. Ava Chen (Cybersecurity Strategist):

Employee Training

Access Control

Communication Protocols

Incident Response

Physical Security

Policy Enforcement

Monitoring Systems

- Michael Rodriguez (IT Infrastructure Architect):

System Redundancies

Database Security

Cloud Infrastructure Security

Hardware Integrity

Software Updates

Server Access Control

Network Segmentation

- Dr. Yara Singh (Organizational Psychologist):

Employee Training Programs

Behavior Analysis

Social Engineering Awareness

Psychological Profiling

Insider Threat Monitoring

Communication Protocols

Staff Vigilance

- Laura García (Vendor Management Specialist):

Vendor Risk Assessment

Third-party Audits

Supply Chain Security

Service Level Agreements

External Collaboration Security

Information Sharing Policies

Outsourcing Management

- Edward Kim (Legal and Compliance Officer):

Regulatory Compliance

Legal Risk Assessment

Policy Development

Contractual Safeguards

Audit Trails

Reporting Mechanisms

Intellectual Property Protection

- Anita Patel (Chief Risk Officer):

Risk Management Framework

Business Continuity

Disaster Recovery Planning

Financial Impact Analysis

Strategic Risk Evaluation

Operational Risk Controls

Compliance Risk Management

- Lt. Col. John Abrams (Physical Security Expert):

Physical Access Controls

Surveillance Systems

Security Personnel Training

Entry Point Security

Emergency Response

Visitor Management

Environmental Controls

# Appendix D

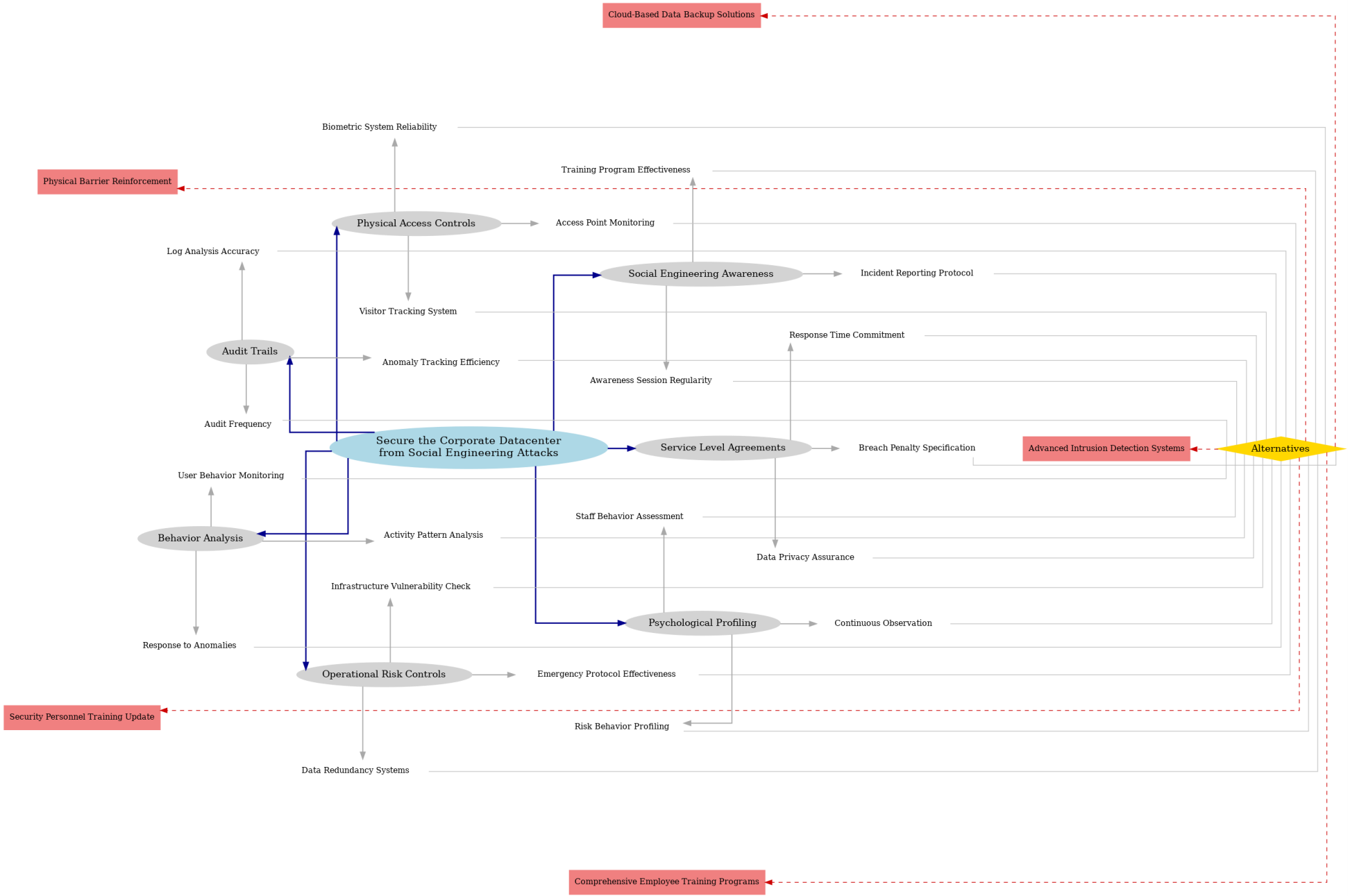

Figure D.1. AHP tree for the goal 'Secure the Corporate Datacenter from Social Engineering Attacks'.